# Clinical gait data analysis based on Spatio-Temporal features

Rohit Katiyar, Lecturer
Computer Science & Engineering Dept.
Harcourt Butler Technological Institute
Kanpur (U.P.), India .

Dr. Vinay Kumar Pathak, Vice Chancellor
Professor, Computer Science & Engg. Dept.
Uttarakhand Open University, Uttarakhand
Haldwani, India .

*Abstract*—**Analysing human gait has found considerable interest in recent computer vision research. So far, however, contributions to this topic exclusively dealt with the tasks of person identification or activity recognition. In this paper, we consider a different application for gait analysis and examine its use as a means of deducing the physical well-being of people. The proposed method is based on transforming the joint motion trajectories using wavelets to extract spatio-temporal features which are then fed as input to a vector quantiser; a self-organising map for classification of walking patterns of individuals with and without pathology. We show that our proposed algorithm is successful in extracting features that successfully discriminate between individuals with and without locomotion impairment.**

## 1. Introduction

### *1.1. Human locomotion*

Normal walking in humans may be defined as a method of locomotion involving the use of two legs, alternately, to provide both support and propulsion, with at least one foot in contact with the ground at all times. Walking is a periodic process and *gait* describes the manner or the style of walking—rather than the walking process itself [1]. Fig. 1 illustrates the repetitive events of gait. The stance phase starts by heel strike (or foot contact with ground) passes through midstance and ends by taking the toe off the ground to start the swing phase. The time interval between two successive occurrences of one of the repetitive events of walking is known as the *gait cycle* and it is usually defined between two consecutive heel-strikes of the same foot. One characteristic phase of walking is the double support interval, i.e. when both feet are in contact with the ground. This time interval decreases as the velocity of the subject increases until it vanishes; the subject is then considered to be running.

The development of photographic methods of recording a series of displacements during locomotion by the end of the 19th century encouraged researchers from different disciplines to study human motion. The images were so useful as studies of the human form in motion that the noted poet and physician Oliver Wendell Holmes, who was interested in providing artificial limbs for veterans of the American Civil War, proclaimed that it was photography which assisted him in the study of the “complex act” of walking [2]. Experiments of the American photographer Eadweard Muybridge of photographing animals (e.g. horse during trotting) and humans in motion (e.g. athletes while practising a sport) perfected the study of animal and human locomotion [3]. Early experiments to study human locomotion were done by Marey, the first scientist in Europe to study motion and its visual implications [4]. Marey observed an actor dressed in a black body stocking with white strips on his limbs. He studied the motion through observing the traces left on photographic plates as the actor walked laterally across the field of view of the camera [5]. Later, at the end of the century, two German scientists Braune and Fischer used a similar approach to study human motion [6], but they used light rods attached to the actor’s limbs instead. Following those pioneers, lots of researchers from different disciplines studied human locomotion. Since the early seventies of the last century, biomechanics researchers have used a technique similar to the ones used by Marey for gait analysis and assessment. In Ref. [7] a method is described for measurement of gait movement from a motion picture film where three cameras were placed such that two are on the sides and one is at the front of a walkway—and a barefooted subject walked across the walkway. Measurements of the flexion/extension of knee and ankle in the sagittal plane and rotation of the pelvis, femur and foot in the transverse plane were measured with this system which had the advantage of being painless and did not involve any encumbering apparatus attached to the patient. In Ref. [8] a television/computer system is designed to estimate the spatial coordinates of markers attached to a subject indicating anatomical landmarks. The system was designed and tested for human locomotion analysis. Another attempt at kinematic analysis using a video camera, frame grabbers and a PC was proposed in Ref. [9]. The approach was based on tracking passive markers attached on specific body landmarks and the results were depicted as an animated stick diagram as well as graphs of joints’ flexion/extension. Interest of researchers was not confined to patterns of normal subjects, it extended to the study of the pathological gait [10, 11].

The method of using markers attached to joints or points of interest of a subject or an articulated object is similar to what is known in the literature of motion perception as *Moving light* *displays (MLDs)*. In Refs. [12, 13], Johansson used MLDs in psychophysical experiments to show that humans can recognize gaits representing different activities such as walking, stair climbing, etc. when watching a sequence of frames of subjects having lights attached to them (sometimes referred to in the literature as Johansson's figures). One experiment had a sequence of 36 motion-picture frames in which two persons were dancing together with 12 lights attached to each one: two at each shoulder, elbow, wrist, hip, knee and ankle. He reported that "naïve" subjects, when shown the sequence, were able to recognise in a fraction of a second that two persons were moving. However, they were not able to identify what a single stationary frame represented. Cutting and Koslowski also showed that using MLDs, one can recognise one's friends [14] and can also determine the gender of a walker [15].

Human motion analysis is a multidisciplinary field which attracts the attention of a wide range of researchers [16]. The nature of the motion analysis research is dictated by the underlying application [17–24]. Motion trajectories are the most widely used features in motion analysis. Most of human motion is periodic, as reflected in changes in joint angle and vertical displacement trajectories, functions involving motion are represented using transformation representing the spatio-temporal characteristics of these trajectories [20] or the volume [25]. In Ref. [26], a computer program that generated absolute motion variables of the human gait from predetermined relative motions was described. Kinematics data during free and forced-speed walking were collected and trajectories were analysed using fast Fourier transform (FFT). It was found that the spectrum of the variables was concentrated in the low frequency range while high frequencies components (above the 15th harmonic) resembled those of white noise.

FFT analysis was also used in Ref. [27]. FFT components of joint displacement trajectories were used as feature vectors to recognise people from their gait. In Ref. [28] the medial axis transformation was used to extract a stick figure model to simulate the lower extremities of the human body under certain conditions. Three male subjects were involved in the study where their 3D kinematic data were averaged to derive a reference sequence for the stick figure model. Two segments of the lower limb (thigh and shank) were modelled and the model was valid only for subjects walking parallel to the image plane. Factors affecting kinematics patterns were explored by studying subjects walking with bare feet and high heels, with folded arms and with arms freely swinging. It was concluded that there was almost no difference in the kinematics patterns. Eigenspace representation was used in Refs. [29–31]. This representation reduced computation of correlation-based comparison between image sequences. In Ref. [29], the proposed template-matching technique was applied for lip reading and gait analysis. The technique was useful in recognising different human gait. In Ref. [30] a combination of eigenspace, transformation and canonical space transformation was used to extract features to recognise six people from their gait.

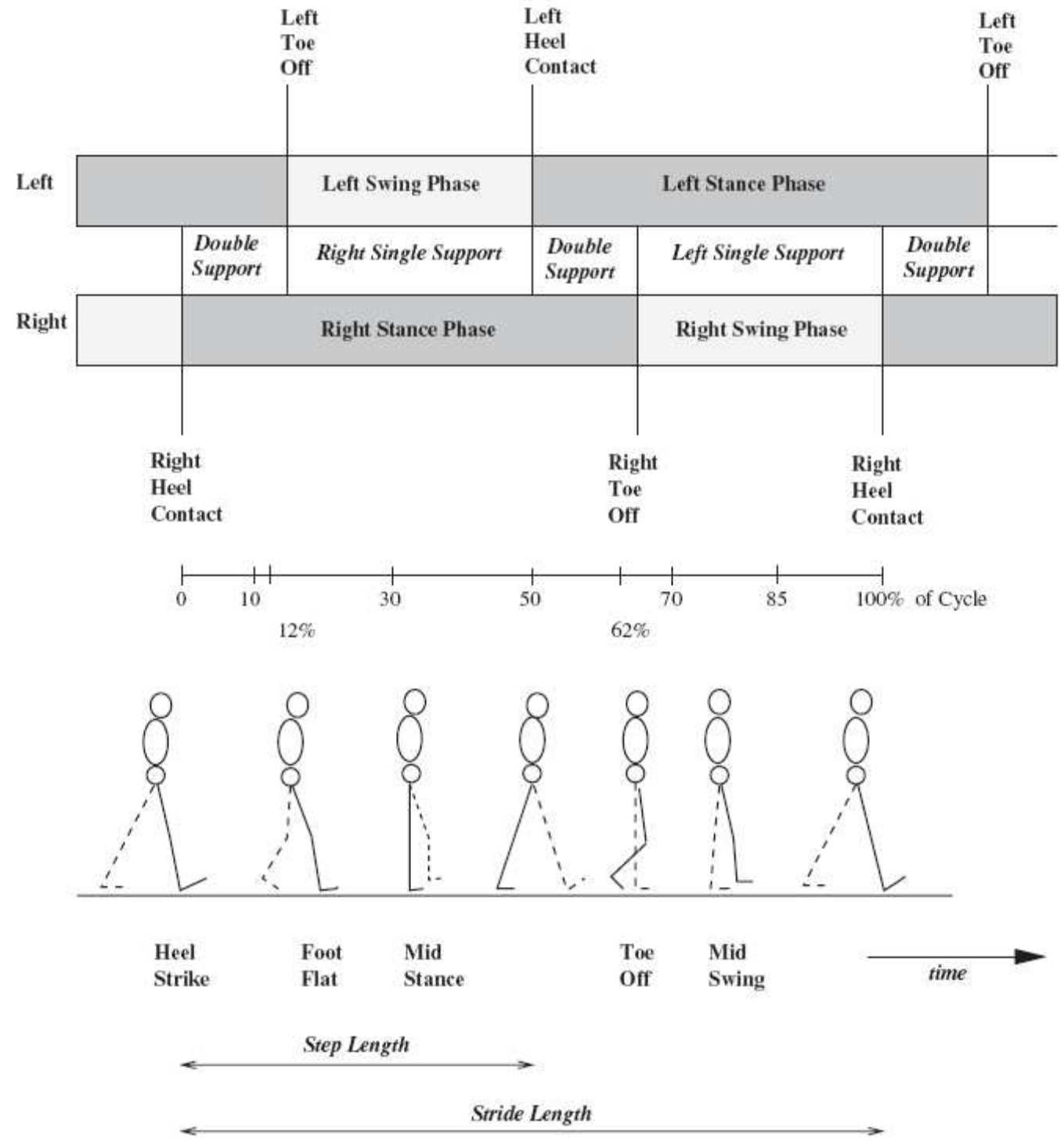

Fig. 1. A typical normal walking cycle illustrating the events of gait.

*1.2. Human motion recognition systems*

The majority of systems implemented for understanding human motion focus on learning, annotation or recognition of a subject or a human movement or activity. For recognising activities, independent of the actor, it is necessary to define a set of unique features that would identify the activity (from other activities) successfully. The authors in Refs. [32–34] presented a general non-structural method for detecting and localising periodic activities such as walking, running, etc. from low-level grey scale image sequences. In their approach, a periodicity measure is defined and associated with the object tracked or the activity in the scene. This measure determined whether or not there was any periodic activity in the scene. A feature vector extracted from a spatio-temporal motion magnitude template was classified by comparing it to reference templates of predefined activity sets. The algorithm tracked a particular subject in a sequence where two subjects were moving. One nice feature of their algorithm was accounting for spatial scale changes in frames, so it was not restricted to motion parallel to the plane of the image. On the other hand, the effect of changing the viewing angle was not addressed in their work. In Ref. [35] a three-level framework for recognition of activities was described in which probabilistic mixture models for segmentation from low-level cluttered video sequences were initially used. Computing spatio-temporal gradient, color similarity and spatial proximity for blobs representing limbs, a hidden Markov model (HMM) was trained for recognising different activities. The training sequences were either tracked MLD sequences or were hand-labelled joints. The Kalman filter used for tracking coped with short occlusions yet some occlusions resulted in misclassification of activity. Sequences seemed to have only one subject in the scene moving parallel to the image plane. In Ref. [36] features from displacements of the body parts in the vertical and horizontal directions were extracted and a classifier based on HMM was used to identify different activities (walking, hopping, running and limping).

A different perspective for recognising activities was portrayed in Ref. [37] by Johansson. His approach focused on high level representations through modelling human recognition of MLDs. He showed that recognition of gait can be achieved through multiresolution feature hierarchies extracted from motion rather than shape information. He applied his approach to recognise three gaits (walking, running and skipping) performed by four different subjects. A similar approach was used in Ref. [38] where techniques based on space curves were developed assuming the availability of 3D Cartesian tracking data to represent movements of ballet dancers. The system learned and recognised nine movements from an un-segmented stream of motion. The idea was based on representing each movement with a set of unique constraints which were extracted from a phase-space that related the independent variables of the body motion. A potential application given by the authors was video annotation for the ever increasing video databases in, for example, entertainment companies and sports teams.

For recognising individuals, examples of research attempts to develop systems to recognise individuals from their gait have been previously discussed. One other attempt [39] computed the optical flow for an image sequence of a walking person and characterised the shape of the motion with a set of sinusoidally varying scalars. Extracting feature vectors composed of the phases of the sinusoids which have shown significant statistical variation, the system was able to discriminate among five subjects. The representation is model-free and only considered subjects walking across the field of view of a stationary camera. Following the above and other studies, gait has been considered as a biometric for individual authentication [40–42]. The idea of the existence of human gait signatures has been widely accepted, e.g. Refs. [43,44]. In this article, we extend this idea to explore the existence of clinical diagnostic signatures from gait data.

### *1.3. Clinical gait analysis*

Normal walking depends on a continual interchange between mobility and stability. Free passive mobility and appropriate muscle action are basic constituents. Any abnormality restricting the normal free mobility of a joint or altering either the timing or intensity of muscle action creates an abnormal gait. Abnormal gait may be due to an injury, disease, pain or problems of motor control. The subject's ability to compensate for the abnormality determines the amount of functionality retained.

However, when these compensations introduce penalties in joint strain, muscle overuse, lack of normal muscle growth or soft tissue contracture, then clinical intervention becomes a necessity. In determining appropriate intervention, gait analysis is used to identify gait defects.

Clinical gait analysis comprises visual assessment, measurement of stride and temporal parameters such as stance, cadence and walking velocity, kinematics dealing with the analysis of joint movements, angles of rotations, etc. and kinetics involving analysis of forces and moments acting on joints and electromyography (EMG) measuring muscle activity [1]. Gait analysis mainly is to document deviations from normal pattern (deviations might be due to habit, pathological reasons or old age), to determine the abnormalities, their severity and their causes, to plan for future treatment which might involve surgery, physiotherapy, or the use of braces, orthosis or any other walking aid, to evaluate the effect of intervention, and finally to measure and assess the change over time with treatment. Gait analysis instrumentation based on infra-red (IR) cameras and computer-aided systems recording the 3D positions of markers attached to the subject has been used to record gait cycles of the patient and produce patterns and plots for clinicians to assess and hence diagnose. To measure kinematics, the subject is filmed while walking using cameras placed on both sides and in front of the walkway such that each marker is seen by at least two cameras at any instant. For kinetics measurements, the subject activates force plates embedded in the walkway. The output of the cameras and the force plates is fed to the computer which estimates the 3D coordinates of the markers and the moments and torques applied on each joint. Kinematics, kinetics measurements and movement trajectories of the joints in the three different planes of movement are plotted for each patient. The pathological traces are plotted together with normal ones to show the variations resulting from the impairment and at the upper left corner of the figure, different gait parameters, e.g. cadence, velocity, etc. are also estimated for the patient. These graphs are then assessed by the specialists. The number of graphs plotted for gait analysis is immense and extracting useful information from such graphs to accomplish the gait analysis objectives mentioned earlier is a demanding and challenging task. This is due to various reasons that include the complexity of the walking process itself, variability of patients' response to treatment, uncertainty in data quality and the difficulty in distinguishing between primary abnormalities and compensations in the gait pattern.

Gait interpretation involves evaluation of all measurements of kinematics, kinetics and EMG to identify abnormalities in gait and hence suggesting and assessing treatment alternatives. The experience of the clinicians' team is the key element for a successful interpretation and this must include the understanding of *normal* gait, efficient and rigorous data collection and adequate data reduction [45]. Early studies by Murray [46,47] aimed to establish ranges of normal values for normal human walking for both men and women from kinematics data analysis. Her studies involved 60 men of 20–65 years of age and 30 women of 20–80 years of age. The main aim of the study was to provide standards concerning speed, stride dimensions as well as angular and linear displacement of the trunk and extremities with which abnormal gait patterns could be compared. Moreover, the study went further trying to find correlations between different parameters, e.g. height and gait parameters; age and displacement patterns.

Developing automatic systems for clinical gait analysis provides objective analysis and interpretation of gait signals. In developing an automatic system for analysis and recognition of gait signals, signal processing not only forms a key element in the analysis extraction and interpretation of information but also plays an important role in the dimensionality reduction [41]. Some artificial intelligence (AI) methods, as artificial neural networks, due to their inherent abilities of generalization, interpolation and fault tolerance offer means to assist in dealing with the challenge of processing huge amounts of data, classifying it through extracting generic diagnostic signatures. A review of the use of these techniques in analysing clinical gait data can be found in Refs. [48,49]. Moreover, psychophysical experiments carried out by Johansson [12,13] and others showed that humans can recognise activities from a sequence of frames containing only points corresponding to specific body landmarks of a subject.

The research presented in this article is motivated by the capabilities of humans to perceive gait from reduced spatiotemporal trajectories, and attempts to give machines a similar gait perception capability. It builds upon Murray's ideas of setting standards for normal walking and investigates the existence of diagnostic signatures that can be extracted from kinematics based features for both normal and pathological subjects. Our objective is to automatically find salient features within trajectories of locomotion from which normal gait standards could be set. Similarly, for abnormal gait, those features could be used for diagnosis of abnormal walking or for establishing relationships among spatio-temporal parameters, gaits and impairment. The long term objective of this work is to provide clinicians with tools for data reduction, feature extraction and gait profile understanding.

## 2. METHODS

### 2.1. Gait data

Experiments involved gait data of 89 subjects with no disabilities of 4–71 years of age and 32 pathological cases of polio, spina-bifida and cerebral palsy (CP) including symmetrical diplegias (dp), left and right asymmetrical (la, ra) diplegias and left and right hemiplegias (lh, rh) were used in our experiments. The data were collected using a *Vicon_* 3D motion capture system at the Anderson Gait Lab in Edinburgh, with markers placed in accordance with Ref. [50]. Temporal and distance parameters of each subject such as cadence, speed, stride length, etc. were calculated. In this work, the focus is concentrated on sagittal angles of the hip and knee joints from which the aim is to extract salient features and diagnostic gait signatures.

### 2.2. Spatio-temporal feature extraction

Kinematics gait signals (e.g. knee flexion/extension trajectory) are non-stationary signals that are rich in dynamic time related data and partly disturbed by noise and artifacts. Extracting generic features as well as specific ones from such a signal implies that the analysis of the signal ought to be done on an adaptable time scale.

The wavelet transform is an effective tool for analysis of non stationary biomedical signals [51]. It provides spectral decomposition of the signal onto a set of basis functions, wavelets, in which representation of temporal dynamics is retained [52]. The continuous wavelet transform (CWT) for a 1D signal *x(t)* is defined as the inner product of the signal function *x(t)* and the wavelet family. This is expressed by the following:

$$W_x^{\psi}(s,\tau) = 1/\sqrt{|s|}\int x(t)\psi^*((t-\tau)/s)\,dt$$

where $\psi^*(t)$ is the conjugate of the transforming function also called the mother wavelet, $\tau$ is the translation parameter, i.e. representing the shift in the time domain, $s(>0)$ is the scale parameter and is proportional to 1/*frequency*.

The transform is convenient for analysing hierarchical structures where it behaves as a mathematical microscope with properties independent of the magnification factor. In other words, low scales correspond to detailed information of a pattern in the signal whereas high scales correspond to global information of the signal while pattern properties are maintained. Given the advantage of providing a spectrum for a signal whilst maintaining its temporal dynamics on an adaptable time scale, we choose to use the CWT to analyze the joint angle trajectories and we use the Morlet wavelet as the mother wavelet. The Morlet wavelet [53] is a locally periodic wave-train. It is obtained by taking a complex sine wave and localising it with a Gaussian envelope as in the following equations:

$$\mathscr{Re}\{g(t)\} = (1/\sqrt{(2\pi)})(\exp -(-t^2/2))\cos(2\pi\nu_0 t) \quad \text{and}$$

$$\mathscr{Im}\{g(t)\} = (1/\sqrt{(2\pi)})(\exp -(-t^2/2))\sin(2\pi\nu_0 t),$$

where $v_o$ is a constant which satisfies the admissibility condition if $v_o > 0.8$.

Once we extract the spatio-temporal features using CWT, the next step in developing an automatic analysing and recognition system is to determine a classification that is most likely to have generated the observed data. Since neither the classes nor their number are defined *a priori*, this is a typical problem of unsupervised classification. We will consider clustering methods to tackle the problem at hand. Clustering methods facilitate the visualisation where one is able to see the groupings of samples which are close together.

Encouraged by extensive literature on data exploration from self-organising maps (SOM), e.g. Refs. [54–56], the SOM is the favored technique for vector quantisation in this analysis. Moreover, the SOM's primary goal is not one of classification but presentation of structures in data, without the need for *a priori* class definitions [57, 58]. This is typical of the case here where we do not want to enforce specified classification based on prior knowledge of the subjects involved in the study. The SOM also exposes new structures that might not be obvious by visual inspection in the feature vectors representing the motion trajectories.

### *2.3. The training algorithm*

The self-organisation process adaptively defines the reference vectors for the neurons, arranges input patterns according to their internal structures and determines the optimal boundaries among the subspaces, hence reflecting the best feature maps representing the input vector patterns. The Kohonen clustering algorithm can be summarised as follows:
Let $\boldsymbol{x} = [x1, x2, \ldots, xn]^{T} \in \mathscr{R}^{n}$ represent the input vector. $\boldsymbol{x}$ is connected to all nodes of the map via reference vectors or weights $\boldsymbol{w}_i$ . Let $t$ represent the epoch number of training and hence $\boldsymbol{w}_i\ (\boldsymbol{t})$ is the weight vector at time $t$. Initialise the weights $\boldsymbol{w}_i$ either to small random values or using available samples of the input $\boldsymbol{x}$. For each epoch $t$,

*Step* 1: Introduce a new input training vector **x** to the network.

*Step* 2: Compute the distance $\boldsymbol{d}_i$ between the input vector pattern $\boldsymbol{x}$ and each reference vector $\boldsymbol{w}_i$ . The distance can be estimated in any chosen metric, the one most commonly used is the Euclidean distance.

*Step* 3: Select the winning reference vector, the index of which $i^{*}$ whose $d^{*}_{i}$ is the minimum and adaptively modify it and the reference vectors in its neighborhood $N_{i^{*}}\ (t)$ as follows:

$$w_i(t+1) = w_i(t) + N_{i*}(t)(x - w_i(t))$$
$$\text{for all } i \in \text{ neighbourhood of } i^{*}(t),$$

where $N_{i^{*}}\ (t)$ is a smooth kernel function defined over the map. For convergence, $N_{i^{*}}\ (t) \rightarrow 0$ when $t \rightarrow \infty$.

*Step* 4: Repeat steps 1–3 until the training set is exhausted and repeat over time $t$ , until the set of weights reaches its final value, i.e. when it is not further adapted.

The Unified Matrix Method (UMM) [59] is a representation of the self-organising map which makes explicit the distances between the neurons of the map. For each neuron, the mean of the distances to its adjacent neurons is calculated and represented as the height of the neuron in the third dimension. The map then is visualised as a number of hills and valleys. Hills represent the borders separating the different classes which are portrayed as valleys. We use this representation to visualise the SOMs after training and testing of the algorithm.

## 3. EXPERIMENTS AND DISCUSSION

The Morlet wavelet used has $v_0 = 1.0$ and the scale range varying between 1 and 25. If the scale is too low, the generated wavelet is too compressed and hence wavelet properties are lost due to under-sampling. On the other hand, if the scale is too high, the wavelet is excessively dilated resulting in extra filtering operations and therefore requires more computation time.

Fig. 2 shows typical scalograms of the Morlet wavelet transform of the sagittal angles of the hip, knee, and ankle joints of normal (top panels) and pathological case (bottom panels). The ordinate represents the frequency ($\alpha$ scale$^{-1}$) and the abscissa is the time; here, time is measured as the percentage of the gait cycle increasing rightwards. Bright areas represent high scalogram values and dark areas low values as indicated by the color bar.

The stance phase constitutes ≈ 60% of the total cycle for normal gait (refer to Fig. 1), hence we choose to split the scalogram vertically at 60% of gait cycle dividing it into stance phase and swing phase. Furthermore, we look for features in different scale levels, either splitting the scalogram horizontally into two sections (levels); high scale (low frequency components) and low scale (high frequency components) separately. Dark color intensities represent low values and light intensities high values. The splitting of the scalogram is shown on the figures by white dotted lines. We therefore have four regions for each scalogram, shown in Fig. 3: (1) stance phase, low scale, (2) swing phase, low scale, (3) swing phase, high scale and (4) stance phase, high scale.

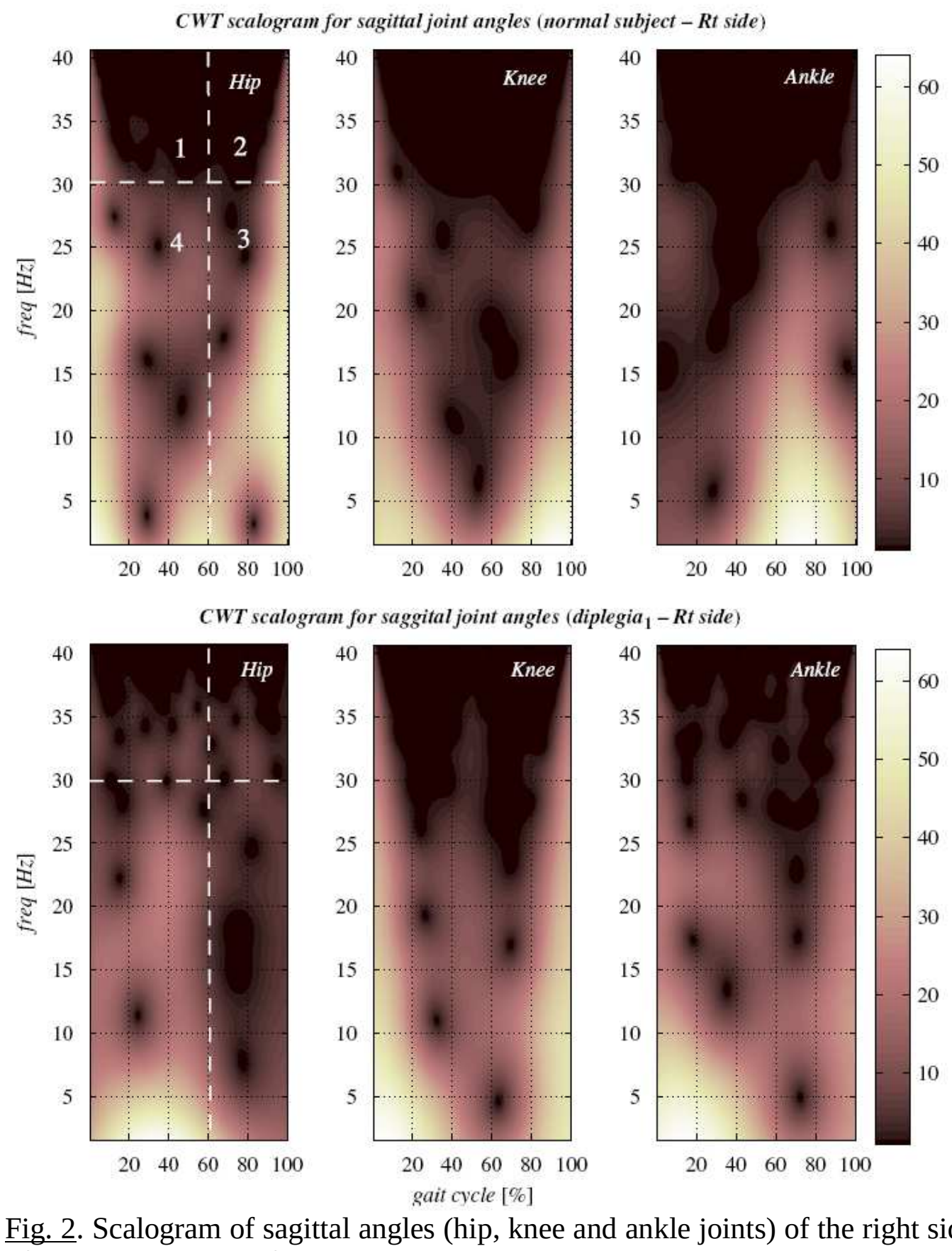

Fig. 2. Scalogram of sagittal angles (hip, knee and ankle joints) of the right side of a normal case (left panels) and a pathological case (right panels). White dotted lines split the scalograms into regions for ease of analysis—see text for explanation.

In this figure, one observes that regions 1 and 2 in hip and knee are mostly dark for a normal subject implying that both joints do not experience high frequency changes in normal gait (remember that frequency $\alpha$ 1/scale). The hip sagittal angle is mainly composed of low frequency components during stance, by mid stance some of those components fade/disappear as represented by the dark spot in region 4 until toe-off (i.e.

beginning of swing phase—region 3) where some of those components start reappearing (bright colour) but with smaller values (darker level shading in region 1). Comparing this normal pattern of the hip to the scalogram of the pathological case (CP) in the bottom panel, we see that in the stance region low frequency components have faded (darker level shading). One also observes existence of high frequency components in both stance and swing phases.

Comparing knee scalograms, one observes in regions 1 and 2 of the CP case that there exist high frequency components even during stance, probably due to spasticity of motion. As for the ankle scalograms, it is evident that in the CP case the ankle joint experiences higher values for high frequency components in regions 1 and 2.

One also observes, in all joint scalograms of the CP case, more discrete dark blobs compared to that of a normal subject which make the scalogram pattern of the normal case more homogeneous with no abrupt changes in shading except when there is a change in gait event (e.g. at toe-off).

In other experiments (results not shown), one observed similarities of patterns between two different normal subjects in all three joint angles in the sagittal plane. Hip and knee scalograms for both subjects exhibited very few high frequency components. Hip comprised low frequency components during stance (extension) which reduce in value during swing (flexion). The knee had small value low frequency components in stance which increase in value during swing. The ankle acquired high valued low frequency at mid-stance and the number of those components increases until toe-off and then decreases to a minimum at heel strike. These features are consistent for a typical normal subject. Any disruption of the sequence of gait events or in the average time taken for any single event results in change in the scalogram pattern. These typical features extracted from the scalograms for a normal subject suggest that automatic classification between normal and pathological cases is possible.

Similarities of scalogram patterns of the right and left sides of a normal subject were also observed showing symmetry in motion of a normal subject whereas obvious discrepancies of the scalogram patterns of the right and the left sides of a pathological subject the asymmetry of motion in all joints. This also means that features extracted from scalograms can be used for testing symmetry/asymmetry of motion in corresponding joints.

When comparing the scalogram of each joint of two different pathological cases, the difference of the patterns representing each pathological case was evident. It is clear by visual inspection of the scalograms that individual with different pathologies generate different signatures. A major typical feature for spastic CP is the high values of low scale components that are represented in the scalogram by dark spots. Typical features for CP hemiplegia and asymmetric diplegia are the high values of scale features at the beginning and end of the gait cycle of the impaired side represented by two concave bright regions at the sides of the scalogram and the asymmetry of the scalograms for all joints. The existence of typical features for different pathologies suggests the possibility of classification based on CWT features.

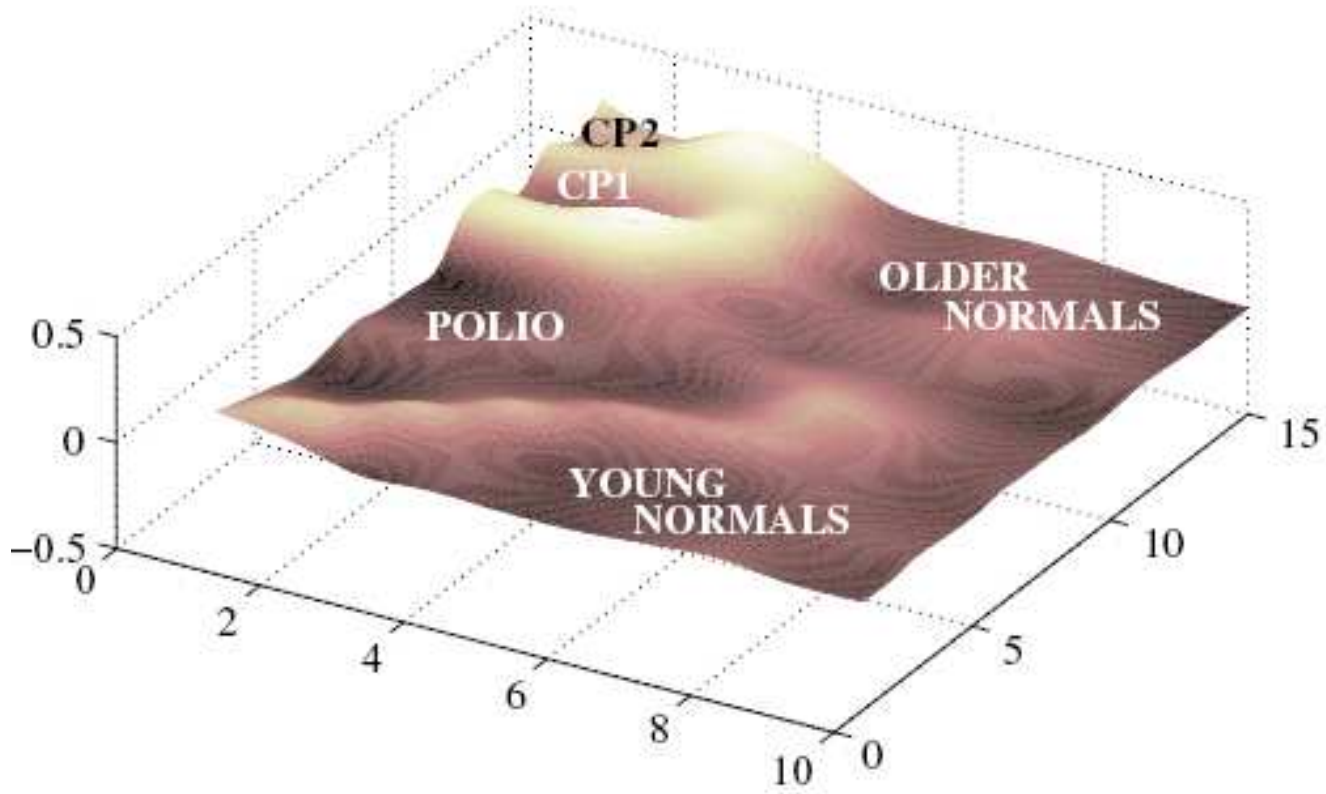

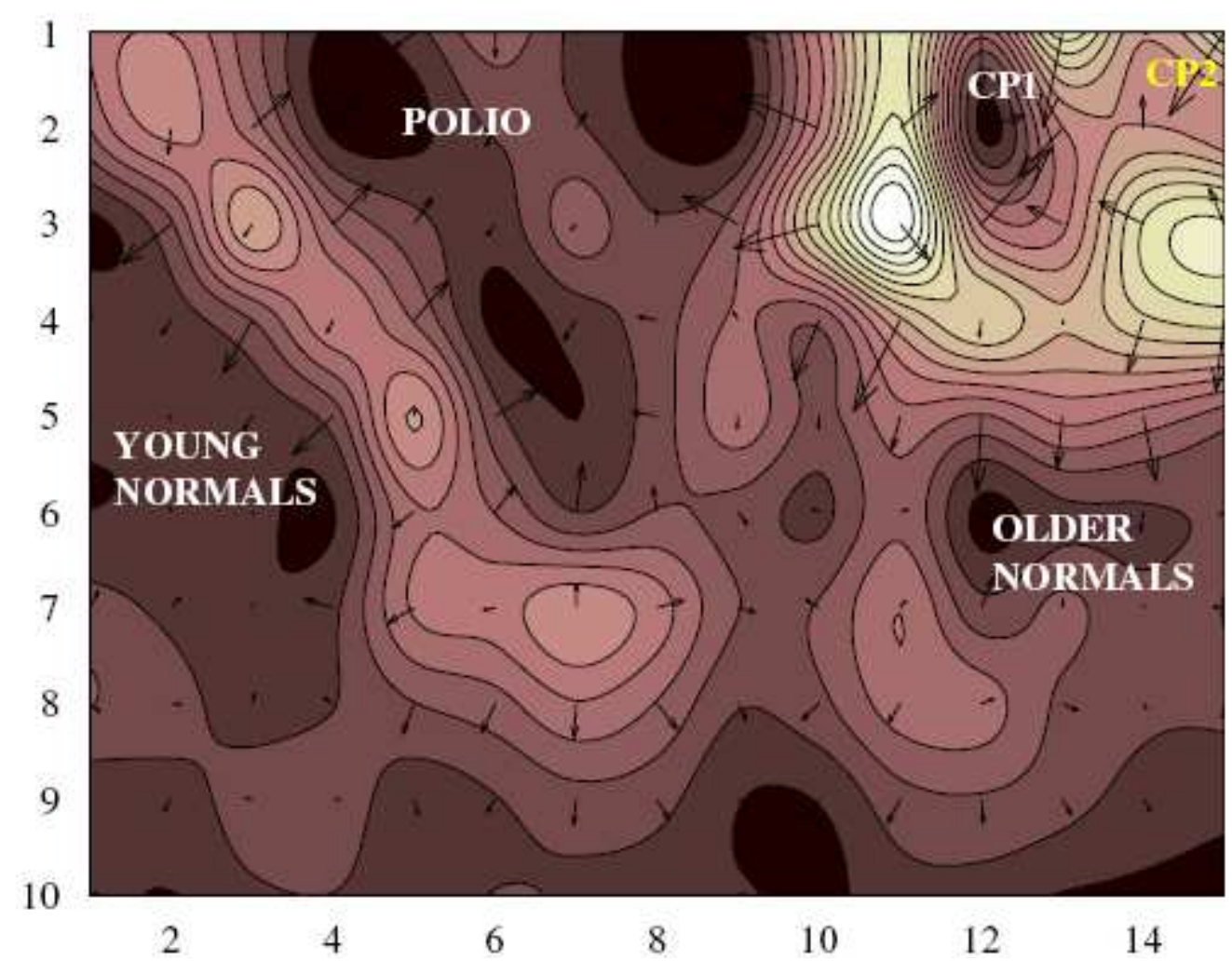

Fig. 3. SOM/UMM (top panel) trained using high scale feature vectors of the sagittal angles of the hip joint of both normal and pathological subject and a contour and quiver diagram (bottom panel) showing the different classes and clustering attraction regions.

The objective of the experiments carried out in this phase was mainly to (1) examine the features that differentiate normal from pathological cases and (2) investigate the existence of diagnostic signatures for different pathologies. The experiments are based only on the sagittal angles, incorporating one or two joints combined together to investigate which joint or combination of joints includes more salient features, and considering one or both sides of the subjects, especially for some pathological cases where there is severe asymmetry. Features are used at two different scale levels, low and high scale features.

The size of the feature vector for a single joint is 20 (time samples, a sample every 5% of the gait cycle) ×8 (scale samples, out of 12) considering one level out of two (high and low).

Extensive experiments were conducted over the gait database described in Section 2 to test the proposed approach.

In all experiments shown in the text, results are illustrated by two plots: (1) a map obtained by applying UMM algorithm to show the different clusters (shown in dark color) separated by the borders (in bright color). The vertical axis in the UMM plot represents the average distance between each node and its neighborhood nodes. The threshold set to discriminate the different clusters depends on the UMM values for every experiment. (2) A quiver and a contour diagram of the map showing the direction of attraction for the different regions.

Both plots are visualisation tools and are used to set the borders of the different clusters.

Due to the inherent ability of abstraction and self organisation, we expect to extract from the trained SOMs the significant features of the different classes. The evaluation of the classifier was performed using leave-one-out validation method on the training data set and using an independent test data set. In the leave-one-out method, one sample is removed from the training data set of size $N$ samples and training is performed using $N-1$ samples and the excluded sample is used for validation. This is repeated $N$ times and classification performance is estimated over all $N$ times. The independent data set (unseen by the classifier in the training phase) is used after training and validation of the map for testing. Classification performance is estimated as the number of misclassified samples over all the number of samples. Kappa coefficient [60] is estimated to assess the agreement above and beyond chance of classification.

*3.1. Differences between normal and pathological subjects*

In order to demonstrate the effectiveness of our proposed approach, we carried out a set of experiments in which we used a data set of 56 subjects including subjects with no disabilities in two different age ranges (20–35 and 55–70 years) and pathologies including CP (diplegics) and poliomyelitis cases. Hip and knee feature vectors are used separately to train SOMs. Data sets were split into two sets of training (40 subjects) and testing (16 subjects).

Fig. 3 shows an example of a map trained with high scale CWT of hip joint feature vectors after applying UMM to visualise the different classes. Hip high scale features showed best results. Classification recognition rate for the training set using leave-one-out validation was 92.5 ± 5% (Kappa = 0.94) and for the independent test set 81.25% (Kappa = 0.9). The SOM successfully discriminates normal from pathological subjects. Furthermore, it recognises the two sets of age ranges separately as well as the different pathological cases based on the global information (high scale) of the hip joint. The quiver diagram shows the direction of attraction for each class of the SOM.

*3.2. Differences between different pathologies*

One technical difficulty of the pathological cases provided is the uniqueness (esp. CP cases) of almost each case, in addition to the complexity of the impairment itself affecting more than one joint as well as the coordination of the overall motion. However, clinicians are sometimes concerned that patients with CP are incorrectly diagnosed as hemiplegic when they are in fact diplegic, which consequently affects their management.

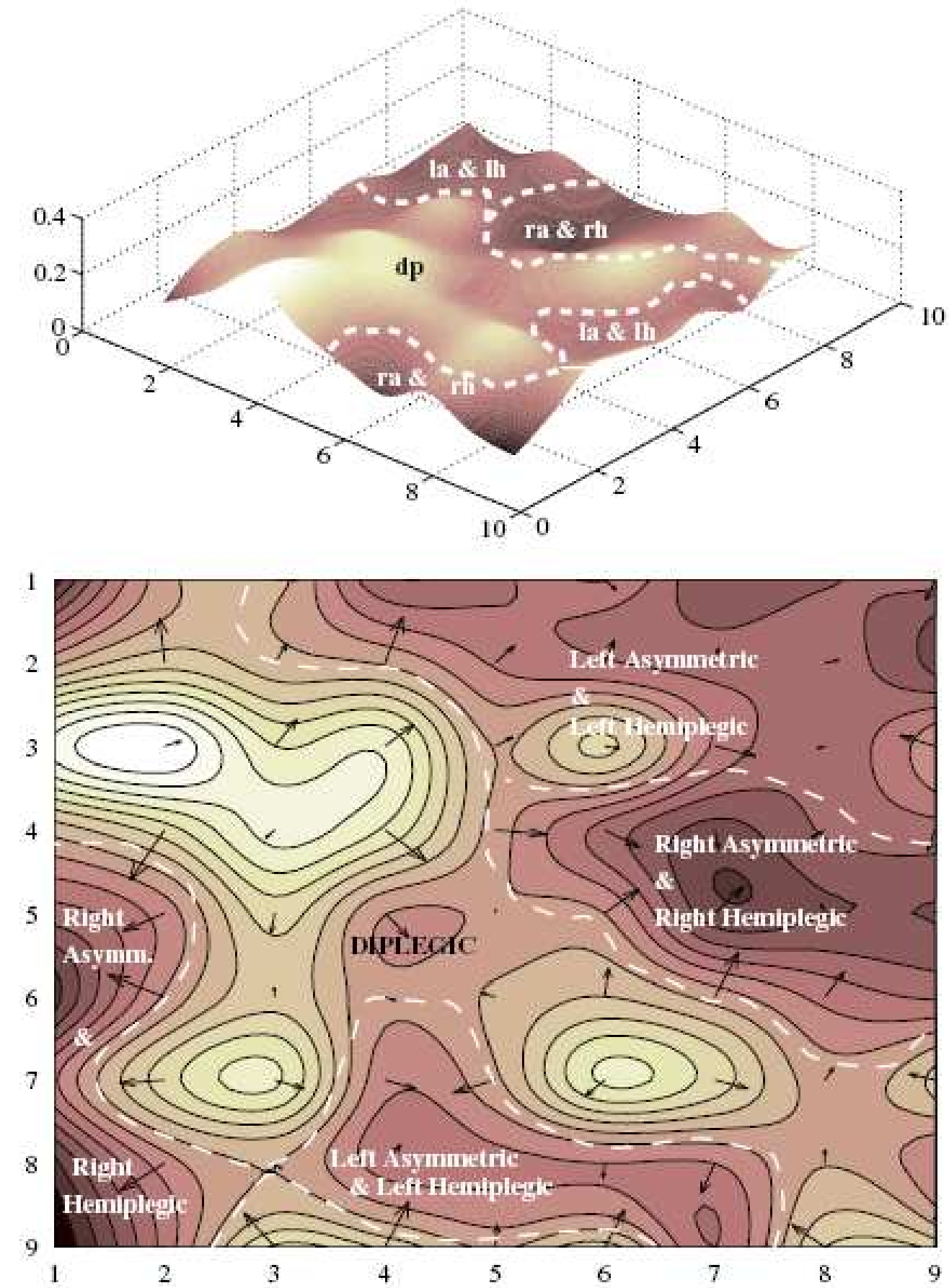

Fig.4. SOM/UMM (top panel) trained using high scale feature vectors of the sagittal angles of combining right and left hip joints of CP pathological subjects and a contour and quiver diagram (bottom panel) showing the different classes and clustering attraction regions.

The objective of this set of experiments is to investigate diagnostic signatures for different pathologies using the proposed algorithm in Section 2.3 for feature extraction and classification. For this set of experiments involving pathological cases only, the same procedure of using leave-one-out cross validation was used for training and validation and an independent data set was used for testing.

Fig. 4 shows the results of classification for subjects with symmetrical diplegias, left and right asymmetrical diplegias and left and right hemiplegias. The map was trained with the high scale features of the hip joint combined together.

The map was self-organised as follows: right asymmetrical diplegias and right hemiplegias clustered together, and so did the left asymmetrical diplegias and left hemiplegias while the symmetrical diplegias grouped between these clusters pulled by their attraction regions in different directions.

Classification recognition rate using leave-one-out for the training data set 91 ± 9% (Kappa = 0.88) and for the test set 70% (Kappa = 0.86). We analysed the results and compared the misclassifications to the clinicians initial diagnosis. These cases were referred to the clinicians for feedback. Table 1 shows the results of three unexpected results visualised in the map of Fig. 4 when testing it and the clinicians interpretations.

Table 1
Analysis of some unexpected results visualised in the map of Fig. 4

| Cases diagnosed before SOM classification | Location in the SOM | Clinical interpretation of the result of the map | Clinical diagnosis after SOM classification |
|---|---|---|---|
| dp-6 | Clustering with lh-4 | dp-6 has upper signs on the left | dp-6 has left hemi as well as diplegia |
| lh-4 | Clustering dp-6 | lh-4 has a pronated and valgus right foot and a catch in his calf muscles | lh-4 is a pretty pure hemi and has very similar patterns to dp-6 at hip and knee on the left. Perhaps also have very mild right sided signs |
| dp-7 | Falls in the same cluster as rh-3 | dp-7 has upper limb signs in the right arm | dp-7 has also hemiplegia on the right |
| rh-3 | Falls in the same cluster as dp-7 | rh-7 has some spasticity in the left foot | rh-3 is in fact an asymmetrical diplegia or diplegia with |
| ra-3 | Falling in the left asymmetric and left hemi cluster | A typographical error by the person providing the data. It is a left asymmetrical case | superimposed hemiplegia ra-3 is a left asymmetrical case |

### 3.3. Discussion

We have carried out a number of experiments to verify our proposed algorithm for feature extraction and classification of gait signatures. The classifier was able to learn and correctly classify almost all samples of the training data set. Good recognition rates were achieved for testing data sets that were not included in training for a range of experiments. The classifier was successful in correctly classifying pathological cases which clinicians, due to the complexity of the impairment, have difficulty in accurately diagnosing. A limitation of SOMs as well as other projection techniques is that the reduction of the dimensionality involves distortions within the map. This might be the reason for some of the misclassifications of the SOM classifier. One suggestion by Ref. [61] is an algorithm called SPINNE where colored graphs are superimposed onto the 2D plots to graphically localize the distortions of the projected data. Applying this algorithm is part of the future work sought for this phase.

The experiments verified that the set of high scale features extracted using CWT discriminates between (1) normal and pathological cases, (2) different pathologies and (3) different groups of CP. However, in general, it was difficult to find consistency in clustering using low scale features due to the wide variations of walks and the high noise level in high frequencies.

The results achieved support our claims that (1) the spatiotemporal analysis performed has maintained the temporal dynamics of the motion trajectories and provided (at least) a similar diagnosis to that of the experts, (2) the classifier was trained only for motion in the sagittal plane and yet resulted in a similar diagnosis to that of the experts which is based on the temporal trajectories in all three planes of motion, at least at a general level of classification. This supports our claim that there is redundant information in the plots which are conventionally used by the clinicians for diagnosis, (3) the proposed classifier permits the display of similar patterns and hence the comparison of cases is made a more interesting and simpler task rather than comparing large number of plots, (4) the classifier demonstrates potential as being a method for clinical gait analysis based on the spectra of the motion trajectories and (5) the spatio-temporal characteristics of motion trajectories are potentially a good candidate to extract diagnostic signatures from gait data.

## 4. Conclusion

In this study, we have investigated the existence of diagnostic signatures based on kinematics gait data for normal and pathological walking.

The work described a method of quantifying generic features of the joint angles of the lower extremities of human subjects in the sagittal plane. The idea is to extract salient diagnostic signatures from hip, knee and ankle joints to characterize normal and pathological walking. The algorithm is based on transforming the trajectories of the flexion/extension of hip and knee and dorsi/plantar flexion of the ankle joints of subjects using the continuous wavelet transform to represent a feature vector which is then fed to a self-organising map (SOM) for clustering. SOM offers a convenient visualisation tool and one can easily see the different clusters after training the map. The clusters were then labelled using the training data set and tested with an independent data set.

The algorithm exhibited its ability to detect and distinguish between normal and pathological subjects, males and females, different age ranges, different pathologies and different categories within a specific pathology (CP).

The procedure correctly classified some difficult pathological cases which were initially misclassified by specialists. This demonstrated its potential as a method of gait analysis and automatic diagnostic signature extraction.